\documentclass[runningheads]{llncs}
\usepackage{graphicx}
\usepackage{multicol}
\usepackage{amsmath}
\usepackage{amssymb}
\usepackage{mathrsfs}
\usepackage{multirow}
\usepackage{cite}
\usepackage{booktabs}
\usepackage{tikz}
\usepackage{xcolor}
\usepackage{caption}
\usepackage{subcaption}
\usepackage{pgfplots}
\usepackage{pgf-umlsd}
\usepackage[]{fp}
\usepackage{xspace}
\usepackage{csquotes}
\usepackage{marvosym}

\usetikzlibrary{calc, positioning, arrows.meta, shapes}

\FPset{totalOffset}{0}
\newcommand{\linkhyper}[3]{
		\draw [#3,line width=0.1mm](#1_front) -- (#2_fll);
		\draw [#3,line width=0.1mm](#1_front) -- (#2_ful);
		\draw [#3,line width=0.1mm](#1_front) -- (#2_bul);
		\draw [#3,line width=0.1mm](#1_front) -- (#2_bll);
}

\newcommand{\trapezium}[8]{

			\def\x{#1} 
			\def\y{#2} 
			\def\wl{#3} 
			\def\wh{#4} 
			\def\h{#5}

	        \coordinate (bl) at  (\x-\wl	,\y-\h);
	        \coordinate (br) at	 (\x+\wl	,\y-\h);
     		\coordinate (tl) at  (\x-\wh	,\y+\h);
         	\coordinate (tr) at  (\x+\wh	,\y+\h);

			\ifthenelse{\equal{#6} {}}
 			{\draw[rotate around={#8:(\x,\y)}](\x-\wl,\y-\h)--(\x+\wl,\y-\h)--(\x+\wh,\y+\h)--(\x-\wh,\y+\h)--cycle;}
			{\draw[#6,fill=#6!20,rotate around={#8:(\x,\y)}](\x-\wl,\y-\h)--(\x+\wl,\y-\h)--(\x+\wh,\y+\h)--(\x-\wh,\y+\h)--cycle;}
			
			\node [font=\fontsize{6}{6}\selectfont,rotate=#8] at (\x,\y) {$\mathrm{#7}$};

}
\newcommand{\networkLayer}[9]{
			\xdef\totalOffset{\totalOffset}
 			\ifthenelse{\equal{#8} {start}}
 			{\FPset{totalOffset}{0}}
 			{}
			\FPeval\currentOffset{(#3)}

			\def\hw{#1} 
			\def\b{0.01}
			\def\c{#2} 
			\def\x{\currentOffset} 
			\def\y{#4} 
			\def\z{#5} 
			\def\inText{#7}

			\coordinate (#8_front)	at	(\x+\c  , \z      , \y);
			\coordinate (#8_back)		at	(\x     , \z      , \y);
			\coordinate (#8_top)		at	(\x+\c/2, \z+\hw/2, \y);
			\coordinate (#8_bottom)	at	(\x+\c/2, \z-\hw/2, \y);
            
			\coordinate (#8_bll) at (0 +\x,  -\hw/2+\z,  -\hw/2+\y); 
			\coordinate (#8_blr) at (\c+\x,  -\hw/2+\z,  -\hw/2+\y); 
			\coordinate (#8_bur) at (\c+\x,   \hw/2+\z,  -\hw/2+\y); 
			\coordinate (#8_bul) at (0 +\x,   \hw/2+\z,  -\hw/2+\y); 
			\coordinate (#8_fll) at (0 +\x,  -\hw/2+\z,   \hw/2+\y); 
			\coordinate (#8_flr) at (\c+\x,  -\hw/2+\z,   \hw/2+\y); 
			\coordinate (#8_fur) at (\c+\x,   \hw/2+\z,   \hw/2+\y); 
			\coordinate (#8_ful) at (0 +\x,   \hw/2+\z,   \hw/2+\y); 

            \ifthenelse{\equal{#9} {}}
 			{}{
 			    \foreach \val in #9
 			        \draw[line width=0.3mm] (\val)--(#8_back);
 			}
 			
			\draw[#6,line width=0.3mm](#8_blr) -- (#8_bur) -- (#8_bul);
			\draw[#6,line width=0.3mm](#8_fll) -- (#8_flr) node[midway,below] {\inText} -- (#8_fur) -- (#8_ful) -- (#8_fll);
			\draw[#6,line width=0.3mm](#8_blr) -- (#8_flr);
			\draw[#6,line width=0.3mm](#8_bur) -- (#8_fur);
			\draw[#6,line width=0.3mm](#8_bul) -- (#8_ful);

			\filldraw[#6,opacity=0.8,line width=0.01mm] ($(#8_fll)+(\b,\b,0)$) -- ($(#8_flr)+(-\b,\b,0)$) -- ($(#8_fur)+(-\b,-\b,0)$) -- ($(#8_ful)+(\b,-\b,0)$) -- ($(#8_fll)+(\b,\b,0)$);
			\filldraw[#6,opacity=0.8,line width=0.01mm] ($(#8_ful)+(\b,0,-\b)$) -- ($(#8_fur)+(-\b,0,-\b)$) -- ($(#8_bur)+(-\b,0,\b)$) -- ($(#8_bul)+(\b,0,\b)$);

			\ifthenelse {\equal{#6} {}}{} 
			{\filldraw[#6,opacity=0.8,line width=0.01mm] ($(#8_flr)+(0,\b,-\b)$) -- ($(#8_blr)+(0,\b,\b)$) -- ($(#8_bur)+(0,-\b,\b)$) -- ($(#8_fur)+(0,-\b,-\b)$);}

}

\begin{document}
\makeatletter
\DeclareRobustCommand\onedot{\futurelet\@let@token\@onedot}
\def\@onedot{\ifx\@let@token.\else.\null\fi\xspace}

\def\eg{\emph{e.g}\onedot} \def\Eg{\emph{E.g}\onedot}
\def\ie{\emph{i.e}\onedot} \def\Ie{\emph{I.e}\onedot}
\def\cf{\emph{c.f}\onedot} \def\Cf{\emph{C.f}\onedot}
\def\etc{\emph{etc}\onedot} \def\vs{\emph{vs}\onedot}
\def\wrt{w.r.t\onedot} \def\dof{d.o.f\onedot}
\def\etal{\emph{et al}\onedot}
\makeatother

\title{Target-aware Bi-Transformer for Few-shot Segmentation\thanks{Supported by Chongqing University.}}
%
%
\author{Xianglin Wang\inst{1}\orcidID{0009-0002-7412-3524} \and
Xiaoliu Luo\inst{1}\orcidID{0000-0002-2365-4950} \and
Taiping Zhang\inst{1}\textsuperscript{(\Letter)}\orcidID{0000-0001-9891-4203}}
\authorrunning{F. Author et al.}
%
\institute{Chongqing University, Chongqing, China 
\email{\{202114131112,20160602026t,tpzhang\}@cqu.edu.cn}}
%
\maketitle              
\begin{abstract}
Traditional semantic segmentation tasks require a large number of labels and are difficult to identify unlearned categories. Few-shot semantic segmentation (FSS) aims to use limited labeled support images to identify the segmentation of new classes of objects, which is very practical in the real world. Previous researches were primarily based on prototypes or correlations. Due to colors, textures, and styles are similar in the same image, we argue that the query image can be regarded as its own support image. In this paper, we proposed the Target-aware Bi-Transformer Network (TBTNet) to equivalent treat of support images and query image. A vigorous Target-aware Transformer Layer (TTL) also be designed to distill correlations and force the model to focus on foreground information. It treats the hypercorrelation as a feature, resulting a significant reduction in the number of feature channels. Benefit from this characteristic, our model is the lightest up to now with only 0.4M learnable parameters. Futhermore, TBTNet converges in only 10\% to 25\% of the training epochs compared to traditional methods. The excellent performance on standard FSS benchmarks of PASCAL-$5^i$ and COCO-$20^i$ proves the efficiency of our method. Extensive ablation studies were also carried out to evaluate the effectiveness of Bi-Transformer architecture and TTL.
\end{abstract}
\section{Introduction}
Semantic segmentation aims to assign each pixel of an image to a certain class, which is one of the cornerstones of computer vision tasks. With the development of the deep convolution network\cite{AlexKrizhevsky2012ImageNetCW,KaimingHe2015DeepRL}, it has made considerable progress. However, the training process requires an enormous amount of labeled data, which is a labor-intensive task. So semi- and weakly-supervised segmentation\cite{YuchaoWangSemiSupervisedSS,XingjiaPan2021UnveilingTP,JungbeomLeeWeaklySS} is invented to reduce the dependence on expensive labels. But all the above methods can only recognize the classes in the train episode. To address this limitation, Few-shot segmentation (FSS) task was proposed.

There are many FSS approaches that have been proposed in recent years\cite{XiaolinZhang2018SGOneSG,YuanweiLiu2022IntermediatePM,JuhongMin2021HypercorrelationSF}. The typical method follows the meta-learning paradigm\cite{AmirrezaShaban2017OneShotLF}, which is easy to overfit due to insufficient data for training. The FSS model is supposed to predict the segmentation of query images based on the condition of support images and corresponding annotations. Nowadays the prevalent approaches are based on prototype\cite{JakeSnell2017PrototypicalNF,YuanweiLiu2022IntermediatePM,LUO2023108811} and pixel-wise correlation\cite{ZhuotaoTian2020PriorGF,JuhongMin2021HypercorrelationSF,SunghwanHong2022CostAW}. Prototype-based methods want to obtain the prototype of the target from support images of high-level features and then utilize the prototype to segment query images. Pixel-wise-based methods take the similarity of each pixel between the query and support images as features to train the model. We can regard similarity as a class-agnostic feature, so the model rarely overfit. 

Objects in different pictures, even if they belong to the same category, may have vastly different features, especially for those parts that are not easily distinguishable. Self-similarity may alleviate this phenomenon, but all the above methods ignored it.

\begin{figure}[t]
\begin{center}
\def\s{1}
\def\imgsize{16}
\resizebox{0.8\linewidth}{!}{
\begin{tikzpicture}
	[scale=\s,
	ArrowC1/.style={rounded corners=\s*1mm,>={Latex[length = 2mm, width = 1.5mm,scale=\s]},line width=\s*1.5pt},
	ArrowC2/.style={rounded corners=\s*0.5mm,>={Latex[length = 1.5mm, width = 1.0mm,scale=\s*0.8]},line width=\s*1.0pt},]
	\tikzstyle{every node}=[scale=\s]
	\node (image_s) at (1,5){\includegraphics[width=\imgsize mm]{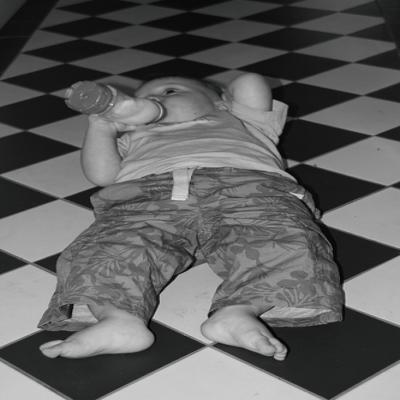}};
	\node (image_q1) at (1,3){\includegraphics[width=\imgsize mm]{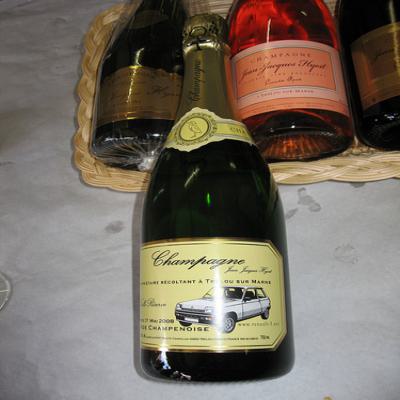}};
	\node (image_q2) at (1,1){\includegraphics[width=\imgsize mm]{figure/split/129_3_fold0.jpg}};
	\node (mask_s) at (5,3.2){\includegraphics[width=\imgsize mm]{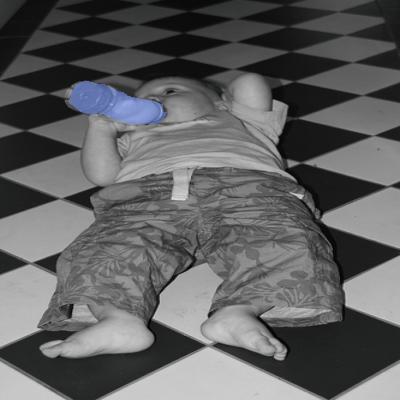}};
	\node [font=\fontsize{7}{7}\selectfont] at (1,6) {Support image};
	\node [font=\fontsize{7}{7}\selectfont] at (1,4) {Query image};
	\node [font=\fontsize{7}{7}\selectfont] at (1,2) {Query image};

	\node (cross) [draw,rectangle,minimum  size=10mm,align=center,scale=0.6] at (3,5){Cross \\ Similarity};
	\node (self) [draw,rectangle,minimum  size=10mm,align=center,scale=0.6] at (3,1){Self \\ Similarity};
	\node (T1) [draw,rectangle,fill=lime!50,minimum  size=12mm,align=center,scale=0.7] at (5,5){Target-aware \\ Transformer};
	\node (C1) [draw,rectangle,fill=yellow!50,minimum  size=5mm,align=center,scale=0.7] at (7.2,5){Conv};
	\node (T2) [draw,rectangle,fill=lime!50,minimum  size=12mm,align=center,scale=0.7] at (5,1){Target-aware \\ Transformer};
	\node (C2) [draw,rectangle,fill=yellow!50,minimum  size=5mm,align=center,scale=0.7] at (7.2,1){Conv};
	\node (mask_q1) at (9,5){\includegraphics[width=\imgsize mm]{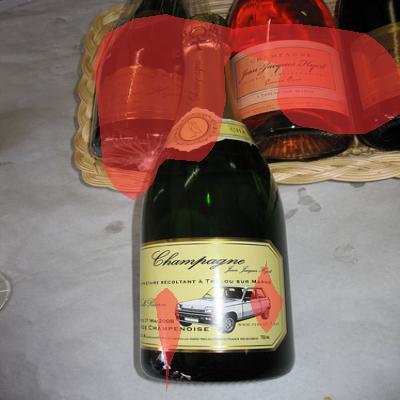}};
	\node (mask_q2) at (9,1){\includegraphics[width=\imgsize mm]{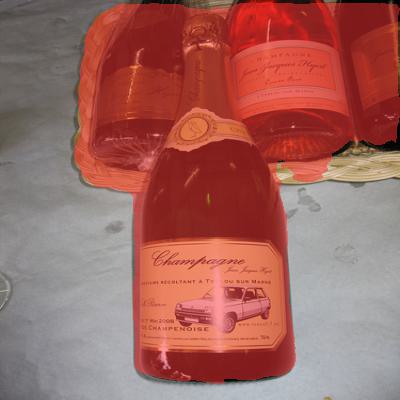}};
	\node [font=\fontsize{7}{7}\selectfont] at (9,6) {Intermediate prediction};
	\node [font=\fontsize{7}{7}\selectfont] at (9,2) {Final prediction};

	\draw [->,ArrowC1,cyan!80] (image_s) -- (cross);
	\draw [->,ArrowC1,cyan!80] (cross) -- (T1);
	\draw [->,ArrowC1,cyan!80] (T1) -- (C1);
	\draw [->,ArrowC1,cyan!80] (C1) -- (mask_q1);
	\draw [->,ArrowC1,cyan!80] ($(image_q1)+(0.9,0.2)$) -| (cross);
	\draw [->,ArrowC1,cyan!80] (mask_s) -- (T1);
	\node [anchor=west,font=\fontsize{7}{7}\selectfont,scale=1.2,cyan!80] at ($(mask_s)+(-0.1,1)$) {Target};
	\node [font=\fontsize{7}{7}\selectfont,rotate=90] at ($(mask_s)+(-1,0)$) {Support mask};

	\node (oplus_1) [font=\fontsize{7}{7}\selectfont,scale=2] at (6.3,1) {$\oplus$};
	\draw [->,ArrowC1,red!80] (image_q2) -- (self);
	\draw [->,ArrowC1,red!80] (self) -- (T2);
	\draw [-,ArrowC1,red!80] (T2) -- ($(oplus_1)+(-0.2,0)$);
	\draw [->,ArrowC1,red!80] ($(oplus_1)+(0.2,0)$) -- (C2);
	\draw [->,ArrowC1,red!80] (C2) -- (mask_q2);
	\draw [->,ArrowC1,red!80] ($(image_q1)+(0.9,-0.2)$) -| (self);
	\draw [->,ArrowC1,red!80] ($(oplus_1)+(0,4)$) -- ($(oplus_1)+(0,0.2)$);
	\draw [ArrowC1,red!80] (mask_q1) -- ++(0,-2.8)[->] -| (T2);
	\node [font=\fontsize{7}{7}\selectfont,red!80,scale=1.2] at (7.2,2.4) {Target};
\end{tikzpicture}
}
\end{center}
\caption{Illustration of the Bi-Transformer architecture for few-shot segmentation.}
\label{fig:long}
\vspace{-10pt}
\end{figure}

In the paper, we proposed the Target-aware Bi-Transformer Network (TBTNet), which can integrate two types of similarity. As shown in Fig. \ref{fig:long}, we first construct an intermediate prediction based on cross-similarity. According to \cite{ZhuotaoTian2020PriorGF}, using high-level feature similarity can make a preliminary prediction, which highlights the most recognizable area for the class. After that we utilize self-similarity to refine the segmentation of query images. It is because that self-similarity contains the structure information of an image, which can expand the part segmentation to the whole.

We also adopt a pyramid structure to implement our model. Our Target-aware Bi-Transformer Module (TBTM) can aggregate two affinity matrices at the same layer, guided by previous layer information, and then transfer the refined similarity to the next layer. That is because high-level intermediate prediction can roughly locate the target, but the boundary of an object is hard to distinguish due to low resolution. Increasing the segmentation resolution progressively with the expansion of affinity matrix can make the boundary more accurate.

In order to make the model only concentrates to the categories we are interested in, we propose a Target-aware Transformer Module (TTM), which consists of two Target-aware Transformer Layers (TTL). Inheriting the virtue of transformer, each pixel of the query image has a global receptive field for the support image. Guided by the mask, TTL only focus on the target in the foreground.
Since our model takes affinity matrices as inputs, which are low-dimensional features, the learned parameters are much less than the vanilla FSS models\cite{KaixinWang2019PANetFI,ZhuotaoTian2020PriorGF,JuhongMin2021HypercorrelationSF,SunghwanHong2022CostAW}. Besides, the training time is shorter, and the computing complexity is lower than other methods due to the fewer parameters. Although our model is small, it can also achieve state-of-the-art performance. All in all, our contributions are:
\begin{itemize}
\item Bi-Transformer architecture is proposed for few-shot segmentation work, which can take the advantage of self-similarity to boost performance.
\item We propose a novel Target-aware Transformer, which can efficiently extract the target's hypercorrelation information under the guidance of the mask.
\item Our TBTNet only has 0.4M learable parameters, it is the lightest FSS model to date.
\item Our model can converge quickly and achieve SOTA performance.
\end{itemize}
\section{Relate works}
\subsection{Few-shot semantic segmentation}
Few-shot semantic segmentation is a branch of semantic segmentation that aims to assign each pixel to a particular class with only a few examples. It was first proposed by \cite{AmirrezaShaban2017OneShotLF}, which adopts a meta-learning paradigm to propagate support branch annotation to query branch. Soon afterward, Jake \etal\cite{JakeSnell2017PrototypicalNF} imports the idea of prototype into FSS, which extracts the prototype from the support set of a certain class and segments the query image by the prototype. Recently, Liu \etal\cite{YuanweiLiu2022IntermediatePM} try to alleviate the intra-class variations, they generate an intermediate prototype from both query and support images.

Although the prototype-based method has had great success in FSS, it disregards a lot of pixel structure information, which hinders the performance of this approach. PFNet\cite{ZhuotaoTian2020PriorGF} uses the affinity matrix between query and support high-level features to obtain a prior segmentation with a parameters-free method. The idea of hypercorrelation was introduced by \cite{JuhongMin2021HypercorrelationSF}, which is a 4D tensor transformed from affinity matrices and squeezed with 4D convolution. Followed \cite{JuhongMin2021HypercorrelationSF}, ASNet\cite{DahyunKangIntegrativeFL} replaced 4D convolution with an attention mechanism based on the transformer to compress hypercorrelation.
\subsection{Vision Transfomer}
Ashish \etal\cite{AshishVaswani2017AttentionIA} first proposed transformer in the Nature Language Processing (NLP) field, which is the standard architecture now. After that, Vit\cite{AlexeyDosovitskiy2020AnII} introduced the transformer to Computer Vision (CV) and achieved great success. Recently, many transformer-based methods have been proposed in FSS. CyCTR\cite{GengweiZhang2021FewShotSV} screens out reliable support features as query tokens to implement cross attention with the query image. DCAMA\cite{XinyuShiDenseCA} aggregates mask by the attention between query and support features.
\section{Problem setting}
There are two sets of data $D_{train}$ and $D_{test}$. The former is used to train the FSS model, and the last one is for testing, to evaluate the accuracy of the model. Each set contains many episodes $E =\{{I^q},{M^q},{I^s},{M^s}\}$ where $I^s$ and $I^q$ represent support image and query image, $M^s$ and $M^q$ denote the corresponding binary mask of the certain category. For the k-shot scenario, $E=\{ {I^q},{M^q},I_1^s,M_1^s,I_2^s,M_2^s,...,I_K^s,M_K^s\}$. The categories of $D_{train}$ and $D_{test}$ are disjoint, which means $C_{train} \cap C_{test} = \mathit{\varnothing}$, where $C_{train}$ and $C_{test}$ are the classes of $D_{train}$ and $D_{test}$. During the training stage, we randomly sample episodes $E$ from $D_{train}$ to learn a network that can predict $M^q$ by $\{ {I^q},I_1^s,M_1^s,I_2^s,M_2^s,...,I_K^s,M_K^s\}$. At the inference stage, our model samples episodes from $D_{test}$ and predicts the novel class target segmentation $M^q$.
\begin{figure*}[t]
\begin{center}
\newcommand{\s}{1}
\newcommand{\FeatureOffset}{2.7}
\newcommand{\simOffset}{3.8}
\newcommand{\HyperOffset}{5.5}
\newcommand{\yOffset}{2.0}
\newcommand{\cmssOffset}{7.5}
\newcommand{\iluOffset}{12.4}
\newcommand{\hThree}{2.3}
\newcommand{\hFour}{3.9}
\resizebox{\textwidth}{!}{
\begin{tikzpicture}
	[scale=\s,
	ArrowC1/.style={rounded corners=\s*1mm,>={Latex[length = 2mm, width = 1.5mm,scale=\s]},line width=\s*1.5pt},
	ArrowC2/.style={rounded corners=\s*0.5mm,>={Latex[length = 1.5mm, width = 1.0mm,scale=\s*0.8]},line width=\s*1.0pt}]

		\begin{scope}[yshift=-4cm]
		\draw [dashed,rounded corners=2mm,line width=0.3mm] (\iluOffset,1.2) rectangle (\iluOffset+5.8,5.0);
		\networkLayer{0.3}{0.1}{\iluOffset+0.5}{0.0}{4.7}{color=red!80}{}{}{}		
		\networkLayer{0.3}{0.1}{\iluOffset+0.5}{0.0}{4.2}{color=blue!50}{}{}{}		

		\networkLayer{0.3}{0.03}{\iluOffset+3.2}{0.0}{4.7}{color=red!40}{}{cql}{}
		\networkLayer{0.3}{0.1}{\iluOffset+3.5}{0.0}{4.7}{color=red!60}{}{csl}{}
		\linkhyper{cql}{csl}{color=red!60}

		\networkLayer{0.3}{0.03}{\iluOffset+3.2}{0.0}{4.2}{color=red!40}{}{sql}{}
		\networkLayer{0.3}{0.1}{\iluOffset+3.5}{0.0}{4.2}{color=blue!30}{}{ssl}{}
		\linkhyper{sql}{ssl}{color=blue!30}

		\networkLayer{0.3}{0.1}{\iluOffset+0.5}{0.0}{3.7}{color=green!30}{}{}{}
		\node at (\iluOffset+3.4,3.7){\includegraphics[width=4 mm]{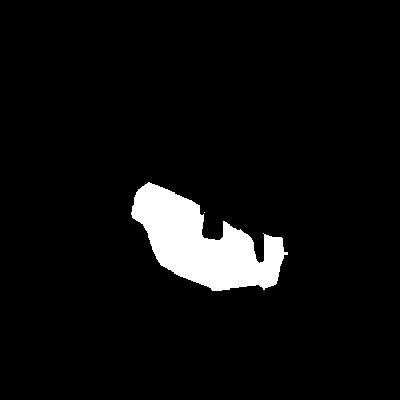}};
		\node at (\iluOffset+0.5,3.2){\includegraphics[width=1.5 mm]{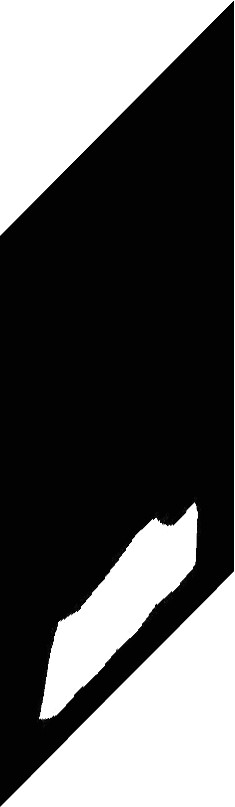}};

		\node (up_l) [draw,fill=white,circle=3mm,font=\fontsize{5}{5}\selectfont] at (\iluOffset+0.5,2.7){};
		\node [font=\fontsize{5}{5}\selectfont] at (up_l) {$\mathrm{up}$};

		\node (TBTM_l) [draw,rectangle,rounded corners=1mm,minimum  height=3mm,minimum  width=8mm,fill=lime!50] at (\iluOffset+0.6,2.2){};
		\node [font=\fontsize{5}{5}\selectfont] at (TBTM_l) {$\mathrm{TBTM}$};
		\node (TTM_l) [draw,rectangle,rounded corners=1mm,minimum  height=3mm,minimum  width=6mm] at (\iluOffset+0.6,1.8){};
		\node [font=\fontsize{5}{5}\selectfont] at (TTM_l) {$\mathrm{TTM}$};
		\node (TTL_l) [draw,rectangle,rounded corners=1mm,minimum  height=3mm,minimum  width=6mm] at (\iluOffset+0.6,1.4){};
		\node [font=\fontsize{5}{5}\selectfont] at (TTL_l) {$\mathrm{TTL}$};

		\node [anchor=west,font=\fontsize{7}{7}\selectfont] at (\iluOffset+1.0,4.7) {Query feature};
		\node [anchor=west,font=\fontsize{7}{7}\selectfont] at (\iluOffset+1.0,4.2) {Support feature};
		\node [anchor=west,font=\fontsize{7}{7}\selectfont] at (\iluOffset+3.7,4.7) {Self-similarity};
		\node [anchor=west,font=\fontsize{7}{7}\selectfont] at (\iluOffset+3.7,4.2) {Cross-similarity};
		\node [anchor=west,font=\fontsize{7}{7}\selectfont] at (\iluOffset+1.0,3.7) {Mix token};
		\node [anchor=west,font=\fontsize{7}{7}\selectfont] at (\iluOffset+3.7,3.7) {Support mask};
		\node [anchor=west,font=\fontsize{7}{7}\selectfont] at (\iluOffset+1.0,3.2) {Predict query mask};
		\node [anchor=west,font=\fontsize{7}{7}\selectfont] at (\iluOffset+1.0,2.7) {$\times 2$ Bi-linear upsampling};
		\node [anchor=west,font=\fontsize{7}{7}\selectfont] at (\iluOffset+1.0,2.2) {Target-aware Bi-Transformer Module};
		\node [anchor=west,font=\fontsize{7}{7}\selectfont] at (\iluOffset+1.0,1.8) {Target-aware Transformer Module};
		\node [anchor=west,font=\fontsize{7}{7}\selectfont] at (\iluOffset+1.0,1.4) {Target-aware Transformer Layer};
\end{scope}
		\node (image_s) at (0.5,3.8){\includegraphics[width=12 mm]{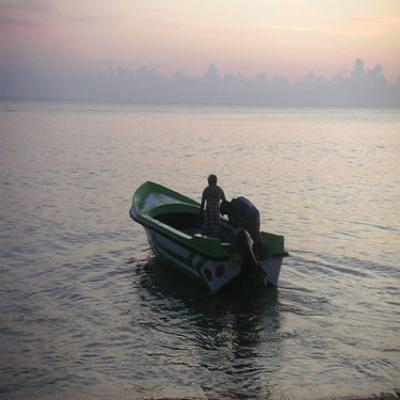}};
		\node (image_q) at (0.5,-1.3){\includegraphics[width=12 mm]{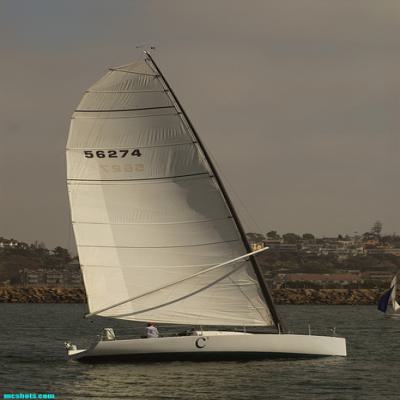}};
		\trapezium{0.5}{2.0}{0.5}{0.8}{0.4}{}{Backbone}{0}
		\trapezium{0.5}{0.5}{0.8}{0.5}{0.4}{}{Backbone}{0}
		\node [font=\fontsize{7}{7}\selectfont] at (0.5,4.7) {Support image};
		\node [font=\fontsize{7}{7}\selectfont] at (0.5,-2.2) {Query image};

		\draw [->,ArrowC1,black!100] (image_s) -- ++(0,-1.4);
		\draw [->,ArrowC1,black!100] (image_q) -- ++(0,1.4);
		\draw [->,ArrowC1,black!100] (0.5,0.9) |- (\FeatureOffset-1.3,1.1);
		\draw [->,ArrowC1,black!100] (0.5,1.6) |- (\FeatureOffset-1.3,1.4);
		\draw [dashed,rounded corners=2mm,line width=0.3mm] (\FeatureOffset-1.3,-2.0) rectangle (\FeatureOffset+0.7,4.2);
		\newcommand{\FeatureSize}{0.3}
		\newcommand{\DimSize}{0.2}
		\newcommand{\xOffset}{0.5}

		\networkLayer{\FeatureSize*4}{\DimSize*0.25}{\FeatureOffset+0.0-\xOffset}{0.0}{0.0}{color=red!80}{}{}{}		
		\networkLayer{\FeatureSize*4}{\DimSize*0.25}{\FeatureOffset+0.2-\xOffset}{0.0}{0.0}{color=red!80}{}{}{}
		\networkLayer{\FeatureSize*4}{\DimSize*0.25}{\FeatureOffset+0.4-\xOffset}{0.0}{0.0}{color=red!80}{}{}{}
		\networkLayer{\FeatureSize*2}{\DimSize*0.5}{\FeatureOffset-0.2-\xOffset}{0.0}{\hThree}{color=red!80}{}{}{}	
		\networkLayer{\FeatureSize*2}{\DimSize*0.5}{\FeatureOffset+0.0-\xOffset}{0.0}{\hThree}{color=red!80}{}{}{}
		\networkLayer{\FeatureSize*2}{\DimSize*0.5}{\FeatureOffset+0.2-\xOffset}{0.0}{\hThree}{color=red!80}{}{}{}
		\networkLayer{\FeatureSize*2}{\DimSize*0.5}{\FeatureOffset+0.4-\xOffset}{0.0}{\hThree}{color=red!80}{}{}{}
		\networkLayer{\FeatureSize}{\DimSize}{\FeatureOffset-0.1-\xOffset}{0.0}{\hFour}{color=red!80}{}{}{}		
		\networkLayer{\FeatureSize}{\DimSize}{\FeatureOffset+0.2-\xOffset}{0.0}{\hFour}{color=red!80}{}{}{}

		\networkLayer{\FeatureSize*4}{\DimSize*0.25}{\FeatureOffset+0.0}{\yOffset}{0.0}{color=blue!50}{}{}{}	
		\networkLayer{\FeatureSize*4}{\DimSize*0.25}{\FeatureOffset+0.2}{\yOffset}{0.0}{color=blue!50}{}{}{}
		\networkLayer{\FeatureSize*4}{\DimSize*0.25}{\FeatureOffset+0.4}{\yOffset}{0.0}{color=blue!50}{}{}{}
		\networkLayer{\FeatureSize*2}{\DimSize*0.5}{\FeatureOffset-0.2}{\yOffset}{\hThree}{color=blue!50}{}{}{}	
		\networkLayer{\FeatureSize*2}{\DimSize*0.5}{\FeatureOffset+0.0}{\yOffset}{\hThree}{color=blue!50}{}{}{}
		\networkLayer{\FeatureSize*2}{\DimSize*0.5}{\FeatureOffset+0.2}{\yOffset}{\hThree}{color=blue!50}{}{}{}
		\networkLayer{\FeatureSize*2}{\DimSize*0.5}{\FeatureOffset+0.4}{\yOffset}{\hThree}{color=blue!50}{}{}{}
		\networkLayer{\FeatureSize}{\DimSize}{\FeatureOffset-0.1}{\yOffset}{\hFour}{color=blue!50}{}{}{}		
		\networkLayer{\FeatureSize}{\DimSize}{\FeatureOffset+0.2}{\yOffset}{\hFour}{color=blue!50}{}{}{}

		\draw [->,ArrowC1,black!100] (\FeatureOffset+0.7,3.5) -- (\HyperOffset-1.2,3.5);
		\draw [->,ArrowC1,black!100] (\FeatureOffset+0.7,1.5) -- (\HyperOffset-1.2,1.5);
		\draw [->,ArrowC1,black!100] (\FeatureOffset+0.7,-0.5) -- (\HyperOffset-1.2,-0.5);

		\node [font=\fontsize{7}{7}\selectfont] at (\FeatureOffset-0.25,4.55) {Feature};
		\node [font=\fontsize{7}{7}\selectfont,scale=0.9] at (\FeatureOffset+0.35,3.5) {$l\!=\!4$};
		\node [font=\fontsize{7}{7}\selectfont,scale=0.9] at (\FeatureOffset+0.35,1.5) {$l\!=\!3$};
		\node [font=\fontsize{7}{7}\selectfont,scale=0.9] at (\FeatureOffset+0.35,-0.5) {$l\!=\!2$};
		\node [font=\fontsize{7}{7}\selectfont,scale=0.6] at (\FeatureOffset-0.3,-1.8) {${{C_l}\!\times\!{H_l}\!\times\!{W_l}}$};
		\node [font=\fontsize{7}{7}\selectfont,scale=0.9] at (\FeatureOffset-0.2,-2.3) {$\{\{{\rm{F}}_{l,d}^s{\rm{,F}}_{l,d}^q\}_{d=1}^{D_l}\}_{l=2}^4$};

		\node [draw,fill=white,rectangle,minimum  height=4mm,minimum  width=50mm,rotate=90] at (\simOffset,1.25){};
		\node [font=\fontsize{6}{6}\selectfont,rotate=90] at (\simOffset,1.25) {$\mathrm{Cosine\ Similarity}$};

		\draw [dashed,rounded corners=2mm,line width=0.3mm] (\HyperOffset-1.2,-2.0) rectangle (\HyperOffset+0.5,4.2);
		\networkLayer{\FeatureSize*4}{0.03}{\HyperOffset-\xOffset}{0.0}{0.0}{color=red!40}{}{cq2}{}		
		\networkLayer{\FeatureSize*2}{0.2}{\HyperOffset}{0.0}{0.0}{color=red!60}{}{cs2}{}
		\linkhyper{cq2}{cs2}{color=red!80}
		\networkLayer{\FeatureSize*2}{0.03}{\HyperOffset-\xOffset}{0.0}{\hThree}{color=red!40}{}{cq3}{}		
		\networkLayer{\FeatureSize*2}{0.3}{\HyperOffset}{0.0}{\hThree}{color=red!60}{}{cs3}{}
		\linkhyper{cq3}{cs3}{color=red!80}
		\networkLayer{\FeatureSize}{0.03}{\HyperOffset-\xOffset}{0.0}{\hFour}{color=red!40}{}{cq4}{}		
		\networkLayer{\FeatureSize}{0.1}{\HyperOffset}{0.0}{\hFour}{color=red!60}{}{cs4}{}
		\linkhyper{cq4}{cs4}{color=red!80}

		\networkLayer{\FeatureSize*4}{0.03}{\HyperOffset}{\yOffset}{0.0}{color=red!40}{}{sq2}{}		
		\networkLayer{\FeatureSize*2}{0.2}{\HyperOffset+0.5}{\yOffset}{0.0}{color=blue!30}{}{ss2}{}
		\linkhyper{sq2}{ss2}{color=blue!30}
		\networkLayer{\FeatureSize*2}{0.03}{\HyperOffset}{\yOffset}{\hThree}{color=red!40}{}{sq3}{}		
		\networkLayer{\FeatureSize*2}{0.3}{\HyperOffset+0.5}{\yOffset}{\hThree}{color=blue!30}{}{ss3}{}
		\linkhyper{sq3}{ss3}{color=blue!30}
		\networkLayer{\FeatureSize}{0.03}{\HyperOffset}{\yOffset}{\hFour}{color=red!40}{}{sq4}{}		
		\networkLayer{\FeatureSize}{0.1}{\HyperOffset+0.5}{\yOffset}{\hFour}{color=blue!30}{}{ss4}{}
		\linkhyper{sq4}{ss4}{color=blue!30}

		\node [font=\fontsize{7}{7}\selectfont] at (\HyperOffset-0.35,4.55) {Hypercorrelation};
		\node [font=\fontsize{7}{7}\selectfont,scale=0.6] at (\HyperOffset-0.35,-1.8) {${{H_l^q}{W_l^q}\!\times\!{H_l^s}{W_l^s}\!\times\!D_l}$};
		\node [font=\fontsize{7}{7}\selectfont,scale=0.9] at (\HyperOffset-0.35,-2.3) {$\{{\rm{X}}_{l}^{qs},{\rm{X}}_{l}^{qq}\}_{l=2}^4$};
		\node (mask_s_4) at (\cmssOffset-1,4.5){\includegraphics[width=6 mm]{figure/split/654_1_fold0.jpg}};
		\node (mask_s_3) at (\cmssOffset-1,2.5){\includegraphics[width=6 mm]{figure/split/654_1_fold0.jpg}};
		\node (mask_s_2) at (\cmssOffset-1,0.5){\includegraphics[width=6 mm]{figure/split/654_1_fold0.jpg}};

		\node [font=\fontsize{7}{7}\selectfont] at ($(mask_s_4)+(-0.1,0.5)$){$M^s$};
		\node [font=\fontsize{7}{7}\selectfont] at ($(mask_s_3)+(-0.1,0.5)$){$M^s$};
		\node [font=\fontsize{7}{7}\selectfont] at ($(mask_s_2)+(-0.1,0.5)$){$M^s$};

		\draw [->,ArrowC1,black!100] (mask_s_4) -| ++(0.7,-0.4);
		\draw [->,ArrowC1,black!100] (mask_s_3) -| ++(0.7,-0.4);
		\draw [->,ArrowC1,black!100] (mask_s_2) -| ++(0.7,-0.4);
		\node (TBTM_4) [draw,rectangle,rounded corners=1mm,minimum  height=10mm,minimum  width=15mm,fill=lime!50] at (\cmssOffset,3.5,0){$\mathrm{TBTM}$};
		\node (TBTM_3) [draw,rectangle,rounded corners=1mm,minimum  height=10mm,minimum  width=15mm,fill=lime!50] at (\cmssOffset,1.5,0){$\mathrm{TBTM}$};
		\node (TBTM_2) [draw,rectangle,rounded corners=1mm,minimum  height=10mm,minimum  width=15mm,fill=lime!50] at (\cmssOffset,-0.5,0){$\mathrm{TBTM}$};

		\draw [<-,ArrowC1,black!100] (TBTM_4) -- ++(-1.5,0);
		\draw [<-,ArrowC1,black!100] (TBTM_3) -- ++(-1.5,0);
		\draw [<-,ArrowC1,black!100] (TBTM_2) -- ++(-1.5,0);

		\draw [->,ArrowC1,black!100] (TBTM_4) -- ++(1.4,0);
		\draw [->,ArrowC1,black!100] (TBTM_3) -- ++(1.3,0);
		\draw [->,ArrowC1,black!100] (TBTM_2) -- ++(1.2,0);

		\draw [dashed,line width=0.3mm] ($(TBTM_4)+(0.7,0.5)$) -- (\iluOffset-0.1,4.8);
		\draw [dashed,line width=0.3mm] ($(TBTM_4)+(0.7,-0.5)$) -- (\iluOffset-0.1,2.0);

		\node (mask_q_4) at (\cmssOffset+1.5,3.5){\includegraphics[width=1.2 mm]{figure/split/3d_mask4.jpg}};
		\node (mask_q_3) at (\cmssOffset+1.5,1.5){\includegraphics[width=2.4 mm]{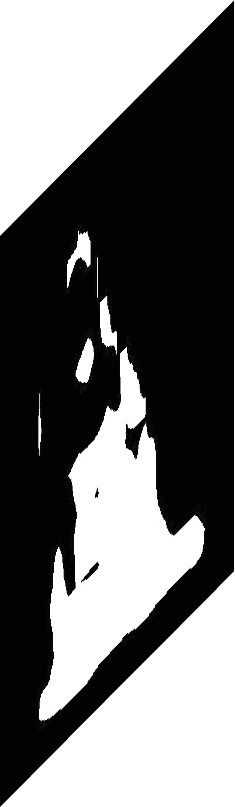}};
		\node (mask_q_2) at (\cmssOffset+1.5,-0.5){\includegraphics[width=4.8 mm]{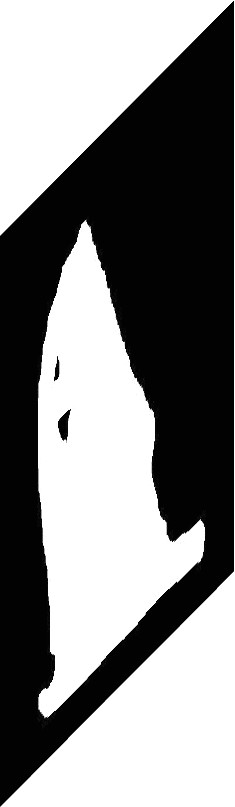}};
		\node (mask_q_1) at (\cmssOffset+4.3,-1.0){\includegraphics[width=9.6 mm]{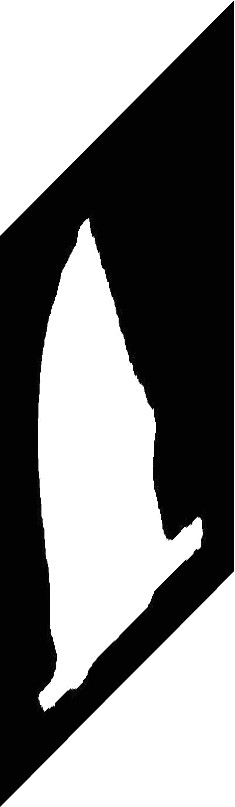}};

		\node [font=\fontsize{7}{7}\selectfont] at ($(mask_q_4)+(-0.1,0.4)$) {$\widehat {\rm{M}}_4^q$};
		\node [font=\fontsize{7}{7}\selectfont] at ($(mask_q_3)+(-0.1,0.6)$) {$\widehat {\rm{M}}_3^q$};
		\node [font=\fontsize{7}{7}\selectfont] at ($(mask_q_2)+(-0.1,0.9)$) {$\widehat {\rm{M}}_2^q$};
		\node [font=\fontsize{7}{7}\selectfont] at ($(mask_q_1)+(-0.1,1.4)$) {$\widehat {\rm{M}}_1^q$};

		\networkLayer{0.3}{0.1}{\cmssOffset}{0.0}{4.7}{color=green!30}{}{t5}{}
		\networkLayer{0.3}{0.1}{\cmssOffset+1.9}{0.0}{3.5}{color=green!30}{}{t4}{}
		\networkLayer{0.6}{0.1}{\cmssOffset+1.9}{0.0}{1.5}{color=green!30}{}{t3}{}
		\networkLayer{1.2}{0.1}{\cmssOffset+2.2}{0.0}{-0.5}{color=green!30}{}{t2}{}

		\node [font=\fontsize{7}{7}\selectfont] at ($(t5_top)+(-0.1,0.2)$) {${\rm{T}}_5$};
		\node [font=\fontsize{7}{7}\selectfont] at ($(t4_top)+(-0.1,0.2)$) {${\rm{T}}_4$};
		\node [font=\fontsize{7}{7}\selectfont] at ($(t3_top)+(-0.1,0.2)$) {${\rm{T}}_3$};
		\node [font=\fontsize{7}{7}\selectfont] at ($(t2_top)+(-0.1,0.2)$) {${\rm{T}}_2$};
        
		\coordinate (intersection) at  ($(TBTM_2)-(-1,1.2)$);
		\draw [->,ArrowC1,black!100] (t5_bottom) -- (TBTM_4);
		\draw [->,ArrowC1,black!100] (TBTM_4) -- (TBTM_3);
		\draw [->,ArrowC1,black!100] (TBTM_3) -- (TBTM_2);
		\draw [ArrowC1,black!100] (TBTM_2) |- (intersection);
		\draw [->,ArrowC1,black!100] ($(TBTM_4)-(0,0.7)$) -| (t4_bottom);
		\draw [->,ArrowC1,black!100] ($(TBTM_3)-(0,0.7)$) -| (t3_bottom);
		\draw [->,ArrowC1,black!100] (intersection) -| (t2_bottom);

		\node (conv_1) [draw,rectangle,rounded corners=1mm,minimum  height=6mm,minimum  width=10mm,scale=0.7] at (\cmssOffset+3,-1.7){$\mathrm{conv}$};
		\node (conv_2) [draw,rectangle,rounded corners=1mm,minimum  height=6mm,minimum  width=10mm,scale=0.7] at (\cmssOffset+3,-0.5){$\mathrm{conv}$};

		\draw [->,ArrowC1,black!100] (intersection) -- (conv_1);
		\draw [->,ArrowC1,black!100] (conv_1) -- (conv_2);
		\draw [->,ArrowC1,black!100] (conv_2) -- ++(0.8,0);

		\node (up_4) [draw,fill=white,circle=3mm,font=\fontsize{5}{5}\selectfont] at (\cmssOffset+3,-1.2){};
		\node [font=\fontsize{5}{5}\selectfont] at (up_4) {$\mathrm{up}$};

		\node (up_4) [draw,fill=white,circle=3mm,font=\fontsize{5}{5}\selectfont] at (\cmssOffset,2.55){};
		\node [font=\fontsize{5}{5}\selectfont] at (up_4) {$\mathrm{up}$};
		\node (up_4) [draw,fill=white,circle=3mm,font=\fontsize{5}{5}\selectfont] at (\cmssOffset,0.55){};
		\node [font=\fontsize{5}{5}\selectfont] at (up_4) {$\mathrm{up}$};


		\newcommand{\iluTBTM}{9.7}
		\draw [dashed,rounded corners=2mm,line width=0.3mm,fill=lime!50] (\iluTBTM,1.1) rectangle (\iluTBTM+8.5,5.1);
		\begin{scope}[xshift=9.7cm,yshift=3.0cm]
	\def\h{1.2}

	\networkLayer{0.6}{0.03}{0.3}{0.0}{\h}{color=red!40}{}{cq1}{}
	\networkLayer{0.6}{0.1}{0.8}{0.0}{\h}{color=blue!30}{}{cs1}{}
	\linkhyper{cq1}{cs1}{color=blue!30}

	\node (mask_s_1) at (1.5,\h+0.5){\includegraphics[width=6 mm]{figure/split/654_1_fold0.jpg}};
	\node (mask_s_2) at (4.5,\h+0.5){\includegraphics[width=6 mm]{figure/split/654_1_fold0.jpg}};
	\node (TTM1) [draw,rectangle,rounded corners=1mm,minimum  height=4mm,minimum  width=8mm,scale=0.7] at (2,\h){$\mathrm{TTM}$};

	\networkLayer{0.6}{0.03}{2.8}{0.0}{\h}{color=red!40}{}{cq2}{}
	\networkLayer{0.3}{0.2}{3.2}{0.0}{\h}{color=blue!30}{}{cs2}{}
	\linkhyper{cq2}{cs2}{blue!30}

	\node (oplus_1) [font=\fontsize{7}{7}\selectfont,scale=2] at (4,\h) {$\oplus$};

	\node (TTM2) [draw,rectangle,rounded corners=1mm,minimum  height=4mm,minimum  width=8mm,scale=0.7] at (5,\h){$\mathrm{TTM}$};

	\networkLayer{0.6}{0.03}{5.8}{0.0}{\h}{color=red!40}{}{cq3}{}
	\networkLayer{0.1}{0.2}{6.0}{0.0}{\h}{color=blue!30}{}{cs3}{}
	\linkhyper{cq3}{cs3}{blue!30}

	\node (conv) [draw,rectangle,rounded corners=1mm,minimum  height=6mm,minimum  width=10mm,scale=0.7] at (7,\h){$\mathrm{conv}$};

	\node (mask_q) at (8,\h){\includegraphics[width=2.8 mm]{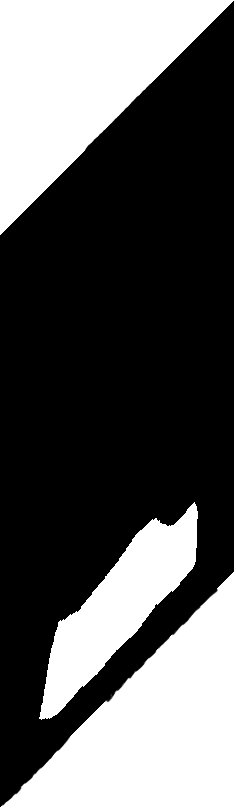}};

	\draw [->,ArrowC1,black!100] (cs1_front) -- (TTM1);
	\draw [->,ArrowC1,black!100] (mask_s_1) -| (TTM1);
	\draw [->,ArrowC1,black!100] (TTM1) -- ++(0.7,0);
	\draw [->,ArrowC1,black!100] (cs2_front) -- ++(0.4,0);
	\draw [->,ArrowC1,black!100] (mask_s_2) -| (TTM2);
	\draw [<-,ArrowC1,black!100] (TTM2) -- ++(-0.8,0);
	\draw [->,ArrowC1,black!100] (TTM2) -- ++(0.7,0);
	\draw [->,ArrowC1,black!100] (cs3_front) -- (conv);
	\draw [->,ArrowC1,black!100] (conv) -- ++(0.8,0);

	\node [anchor=west,font=\fontsize{7}{7}\selectfont,scale=0.7] at ($(cq1_bottom)+(-0.2,-0.25)$) {${{H_l^q}\!{W_l^q}\!\times\!{H_l^s}\!{W_l^s}\!\times\!D_l}$};
	\node [anchor=west,font=\fontsize{7}{7}\selectfont,scale=0.8] at ($(cq1_top)+(0.1,0.2)$) {${\rm{X}}_l^{qs}$};
	\node [anchor=west,font=\fontsize{7}{7}\selectfont,scale=0.7] at ($(cq2_bottom)+(-0.2,-0.25)$) {${{H_l^q}\!{W_l^q}\!\times\!{\widetilde{H}_l^s}\!{\widetilde{W}_l^s}\!\times\!D}$};
	\node [anchor=west,font=\fontsize{7}{7}\selectfont,scale=0.8] at ($(cq2_top)+(0.1,0.2)$) {${\rm\widetilde{X}}_l^{qs}$};
	\node [anchor=west,font=\fontsize{7}{7}\selectfont,scale=0.7] at ($(cq3_bottom)+(0.1,-0.1)$) {${{H_l^q}\!{W_l^q}\!\times\!1\!\times\!D}$};
	\node [anchor=west,font=\fontsize{7}{7}\selectfont,scale=0.8] at ($(cq3_top)+(0.1,0.0)$) {${\rm\overline{X}}_l^{qs}$};
	\node [font=\fontsize{7}{7}\selectfont,scale=0.7] at ($(mask_q)+(0.0,-0.6)$) {${{H_l^q}\!{W_l^q}\!\times\!2}$};
	\node [anchor=west,font=\fontsize{7}{7}\selectfont,scale=0.8] at ($(mask_q)+(-0.4,0.5)$) {${\rm\widehat{M}}_{l}^q$};
	\networkLayer{0.6}{0.3}{0.3}{0.0}{0}{color=green!30}{}{}{}

	\networkLayer{0.6}{0.03}{1.2}{0.0}{0}{color=red!40}{}{cqt}{}
	\networkLayer{0.1}{0.2}{1.4}{0.0}{0}{color=green!30}{}{cst}{}
	\linkhyper{cqt}{cst}{green!30}

	\node [font=\fontsize{7}{7}\selectfont] at ($(cqt_back)-(0.3,0)$) {$=$};
	\node [font=\fontsize{7}{7}\selectfont,scale=2] at (5.8,0) {$+$};
	\networkLayer{0.6}{0.3}{7.9}{0.0}{0}{color=green!60}{}{tl}{}

	\node [anchor=west,font=\fontsize{7}{7}\selectfont,scale=0.7] at ($(cqt_bottom)+(0.1,0.05)$) {${{H_l^q}\!{W_l^q}\!\times\!1\!\times\!D}$};
	\node [anchor=west,font=\fontsize{7}{7}\selectfont,scale=0.8] at ($(cqt_top)+(0.1,-0.0)$) {${\rm{T}}_{l+1}$};
	\node [font=\fontsize{7}{7}\selectfont,scale=0.7] at ($(tl_bottom)+(0.0,-0.25)$) {${{H_l^q}\!{W_l^q}\!\times\!D}$};
	\node [anchor=west,font=\fontsize{7}{7}\selectfont,scale=0.8] at ($(tl_top)+(-0.5,0.0)$) {${\rm{T}}_{l}$};
	
	\draw [->,ArrowC1,black!100] (cst_front) -- ++(4,0);
	\draw [->,ArrowC1,black!100,rounded corners=\s*3mm] (cst_front) -| (4,1);
	\draw [->,ArrowC1,black!100,rounded corners=\s*3mm] (cst_front) -| (4,-1);
	\draw [->,ArrowC1,black!100] (6,0) -- ++(1.8,0);
	\networkLayer{0.6}{0.03}{0.3}{0.0}{-\h}{color=red!40}{}{sq1}{}
	\networkLayer{0.6}{0.1}{0.8}{0.0}{-\h}{color=red!60}{}{ss1}{}
	\linkhyper{sq1}{ss1}{color=red!60}

	\node (mask_q_1) at (1.5,-0.7){\includegraphics[width=6 mm]{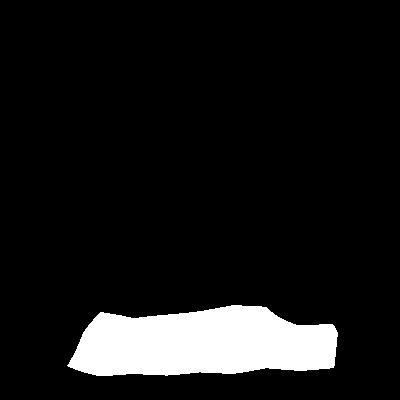}};
	\node (mask_q_2) at (4.5,-0.7){\includegraphics[width=6 mm]{figure/split/654_6_fold0.jpg}};
	\node (TTM3) [draw,rectangle,rounded corners=1mm,minimum  height=4mm,minimum  width=8mm,scale=0.7] at (2,-\h){$\mathrm{TTM}$};

	\networkLayer{0.6}{0.03}{2.8}{0.0}{-\h}{color=red!40}{}{sq2}{}
	\networkLayer{0.3}{0.2}{3.2}{0.0}{-\h}{color=red!60}{}{ss2}{}
	\linkhyper{sq2}{ss2}{red!60}

	\node [font=\fontsize{7}{7}\selectfont,scale=2] at (4,-\h) {$\oplus$};

	\node (TTM4) [draw,rectangle,rounded corners=1mm,minimum  height=4mm,minimum  width=8mm,scale=0.7] at (5,-\h){$\mathrm{TTM}$};

	\draw [dashed,line width=0.3mm] ($(TTM4)+(0.35,0.1)$) -- (6.7,-0.7);
	\draw [dashed,line width=0.3mm] ($(TTM4)+(0.35,-0.1)$) -- (6.7,-1.4);

	\networkLayer{0.6}{0.03}{5.8}{0.0}{-\h}{color=red!40}{}{sq3}{}
	\networkLayer{0.1}{0.2}{6.0}{0.0}{-\h}{color=red!60}{}{ss3}{}
	\linkhyper{sq3}{ss3}{red!60}

	\draw [->,ArrowC1,black!100] (ss1_front) -- (TTM3);
	\draw [->,ArrowC1,black!100] (mask_q_1) -| (TTM3);
	\draw [->,ArrowC1,black!100] (TTM3) -- ++(0.7,0);
	\draw [->,ArrowC1,black!100] (ss2_front) -- ++(0.4,0);
	\draw [->,ArrowC1,black!100] (mask_q_2) -| (TTM4);
	\draw [<-,ArrowC1,black!100] (TTM4) -- ++(-0.8,0);
	\draw [->,ArrowC1,black!100] (TTM4) -- ++(0.7,0);

	\node [anchor=west,font=\fontsize{7}{7}\selectfont,scale=0.7] at ($(sq1_bottom)+(-0.2,-0.25)$) {${{H_l^q}\!{W_l^q}\!\times\!{H_l^s}\!{W_l^s}\!\times\!D_l}$};
	\node [anchor=west,font=\fontsize{7}{7}\selectfont,scale=0.8] at ($(sq1_top)+(0.1,0.2)$) {${\rm{X}}_l^{qq}$};
	\node [anchor=west,font=\fontsize{7}{7}\selectfont,scale=0.7] at ($(sq2_bottom)+(-0.2,-0.25)$) {${{H_l^q}\!{W_l^q}\!\times\!{\widetilde{H}_l^s}\!{\widetilde{W}_l^s}\!\times\!D}$};
	\node [anchor=west,font=\fontsize{7}{7}\selectfont,scale=0.8] at ($(sq2_top)+(0.1,0.2)$) {${\rm\widetilde{X}}_l^{qq}$};
	\node [anchor=west,font=\fontsize{7}{7}\selectfont,scale=0.7] at ($(sq3_bottom)+(0.1,-0.1)$) {${{H_l^q}\!{W_l^q}\!\times\!1\!\times\!D}$};
	\node [anchor=west,font=\fontsize{7}{7}\selectfont,scale=0.8] at ($(sq3_top)+(0.1,-0.1)$) {${\rm\overline{X}}_l^{qq}$};

	\draw [->,ArrowC1,black!100] (cq3_bottom) -- ++(0,-0.7);
	\draw [->,ArrowC1,black!100] (sq3_top) -- ++(0,0.7);

	\def\iluOffset{6.5}
	\def\blockOffset{0.3}
	\draw [dashed,rounded corners=2mm,line width=0.3mm] (6.7,-1.45) rectangle (7.9,-0.6);
	\node (mask) [draw,rectangle,minimum  size=3mm,font=\fontsize{5}{5}\selectfont] at (\iluOffset+\blockOffset+0.3,-0.35){};
	\node [font=\fontsize{6}{6}\selectfont] at (mask) {M};
	\draw [->,ArrowC1,black!100] (\iluOffset,-1.1) -- (\iluOffset+1.8,-1.1);


	\node (ftl_1) [draw,fill=lime!50,rectangle,minimum  height=3mm,minimum  width=6mm,rotate=90] at (\iluOffset+\blockOffset+0.3,-1.1){};
	\node [font=\fontsize{6}{6}\selectfont,rotate=90] at (\iluOffset+\blockOffset+0.3,-1.1) {TTL};
	\node (ftl_2) [draw,fill=lime!50,rectangle,minimum  height=3mm,minimum  width=6mm,rotate=90] at (\iluOffset+\blockOffset+0.7,-1.1){};
	\node [font=\fontsize{6}{6}\selectfont,rotate=90] at (\iluOffset+\blockOffset+0.7,-1.1) {TTL};

	\draw [->,ArrowC2,black!100] (mask) -- (ftl_1);
	\draw [->,ArrowC2,black!100] (mask) -| (ftl_2);
		\end{scope}
	\end{tikzpicture}
}
\end{center}
\caption{Overall network architecture. Our TBTNet consists of four main sub-modules: feature extraction, similarity computation, TBTM pyramidal encoder, and a simple convolution decoder. For more details please refer to Sec.\ref{Method}.}
\label{fig:model}
\vspace{-10pt}
\end{figure*}
\section{Method}
\label{Method}
\subsection{Overview}
As shown in Fig. \ref{fig:model}, our Target-aware Bi-Transformer Network (TBTNet) consists of three Target-aware Bi-Transformer Modules and two decoders. Firstly, a pre-trained backbone is used to extract the features of support image and query image respectively. After that, computing the cosine similarity between query and support, query and itself for hypercorrelation. Next, the hypercorrelation with the same resolution will be mixed by a TBTM. The output of TBTM contains a predicted query mask and a tensor which is the input of the next TBTM. The output from the last TBTM mixed all the hypercorrelation information and will be sent to decoder to make the final prediction.
\subsection{Hypercorrelation Features Computation}

Following\cite{JuhongMin2021HypercorrelationSF}, we take ResNet50 and ResNet101\cite{KaimingHe2015DeepRL}, which is pre-trained on ImageNet\cite{AlexKrizhevsky2012ImageNetCW}, as the backbone to extract features of images. ${\rm{F}}_{l,d}^s{\rm{,F}}_{l,d}^q \in {\mathbb{R}^{{C_l} \times {H_l} \times {W_l}}}$ are the features of support and query images ${{I}}^s,{{I}}^q \in {\mathbb{R}^{{3} \times {H} \times {W}}}$ respectively.
\begin{equation}
{\rm{\{ \{ F}}_{l,d}^\ast {\rm{\} }}_{d = 1}^{{D_l}}{\rm{\} }}_{l = 2}^4{\rm{ = ResNet(}}{I^\ast}),
\end{equation}
where $l$ denotes the output layer of ResNet, $D_l$ means the number of block at layer $l$ and $\ast \in \{s,q\}$. 
\subsubsection{Cross- \& self- similarity.} Since the features extracted from the backbone contain rich semantic information, we compute the cosine between features and get cross-similarity ${\rm{A}}_{l,d}^{qs} \in {\mathbb{R}^{{H_l^q} \, {W_l^q} \times {H_l^s} \, {W_l^s}}}$:
\begin{equation}
{\rm{A}}_{l,d}^{qs}(p^q,p^s)={\rm{ReLU}}(\frac{{\rm{F}}_{l,d}^q (p^q)^T \, {\rm{F}}_{l,d}^s (p^s)}{\Vert {\rm{F}}_{l,d}^s (p^q) \Vert \Vert{\rm{F}}_{l,d}^q (p^s) \Vert}),
\end{equation}
where $p^\ast$ is 2D positions and ${\rm{F}}_{l,d}^\ast (p^\ast) \in \mathbb{R}^{C_l\times1}$. 
We compute the self-similarity ${\rm{A}}_{l,d}^{qq}$ in the same way as cross-similarity, only replacing ${\rm{F}}_{l,d}^s$ with ${\rm{F}}_{l,d}^q$.

\subsubsection{Hypercorrelation.} To obtain cross-hypercorrelation ${\rm{X}}_{l}^{qs}$ and self-hypercorrelation ${\rm{X}}_{l}^{qq} \in {\mathbb{R}^{{H_l^q} \, {W_l^q} \times {H_l^s} \, {W_l^s} \times D_l}}$, we stack all the affinity matrix at the same layer.
\begin{equation}
{\rm{X}}_l^{q\ast}{\rm{  =  Stack(\{ A}}_{l,d}^{q\ast}\} _{d = 1}^{{D_l}}),
\end{equation}
\subsection{Target-aware Bi-Transformer Module}

Previous approaches only use cross-similarity to predict the segmentation of query image, often leading to incomplete result. It greatly limits the capability of the model. In contrast, self-similarity contains the structural information inherent in the image, which helps make the prediction more complete. Therefore, we designed the TBTM, which first  passes the cross-hypercorrelation through two Target-aware Transformer Modules (TTM) and a Convolution Block to obtain the intermediate prediction, then, under the guidance of the prediction, refines the self-hypercorrelation through the other two TTMs. As shown in Fig. \ref{fig:model}, TTM aims to reduce the support spatial sizes progressively and change the channels of the hypercorrelation.
\begin{equation}
\widetilde {\rm{X}}_l^{qs}{\rm{  =  TTM(X}}_l^{qs}{\rm{,}}{M^s}{\rm{)}},
\end{equation}
\begin{equation}
\overline {\rm{X}} _l^{qs}{\rm{  =  TTM(}}\widetilde {\rm{X}}_l^{qs} \oplus {{\rm{T}}_{l + 1}}{\rm{,}}{M^s}{\rm{)}},
\end{equation}
where ${M^s \in \{0,1\}^{H \times W}}$ is the binary segmentation map of support image. $\widetilde {\rm{X}}_l^{qs} \in {\mathbb{R}^{{H_l^q} \, {W_l^q} \times {{\widetilde H}_l^s} \, {{\widetilde W}_l^s} \times D}}$ and $\overline {\rm{X}}_l^{qs} \in {\mathbb{R}^{{H_l^q} \, {W_l^q} \times 1 \,  \times D}}$ are the output of TTM. Noting that both ${{\widetilde H}_l^s}$ and ${{\widetilde W}_l^s}$ are smaller than $H_l^s$ and $W_l^s$ respectively. ${{\rm{T}}_{l + 1}}$ denotes the MixToken from previous layer which has mixed the self- and cross-similarity information. It is initialized to 0 and ${{\rm{T}}_{l + 1}} \in {\mathbb{R}^{{H_l^q} \, {W_l^q} \times 1 \,  \times D}}$. We utilized broadcasted element-wise addition  to sum $\widetilde {\rm{X}}_l^{qs}$ and ${{\rm{T}}_{l + 1}}$ because of their shapes are different. $\overline {\rm{X}} _l^{qs}$ will be sent to a convolution block to compute $\widehat {\rm{M}} _l^q$:
\begin{equation}
\widehat {\rm{M}}_l^q = {\rm{ReLU(Conv(ReLU(Conv(\overline {X}}} _l^{qs})))),
\end{equation}
where $\widehat {\rm{M}}_l^q$ denotes the predicted segmentation of query image at layer $l$, $\widehat {\rm{M}}_l^q \in \mathbb{R}^{2 \times H_l^q \times W_l^q}$. The convolution block consists of two times alternating convolution layers and ReLU activation functions. We can get a binary segmentation map ${\rm{M}}_l^q \in \{0,1\}^{H_l^q \times W_l^q}$ easily from $\widehat {\rm{M}}_l^q$:
\begin{equation}
{\rm{M}}_l^q(x,y) = \left\{ \begin{array}{lll}
{\rm{0}}& {\rm{if}} &\widehat {\rm{M}}_l^q(0,x,y) > \widehat {\rm{M}}_l^q(1,x,y)\\
1 & \multicolumn{2}{c}{\rm{otherwise}}
\end{array}\right.,
\end{equation}
where $x \in \{0,1,...,W_l^q-1\},y \in \{0,1,...,H_l^q-1\}$ indicate the 2D coordinates of features. And then, we can treat ${\rm{M}}_l^q$ as a pseudo mask and deal with the self-similarity ${\rm{X}}_l^{qq}$ as same as ${\rm{X}}_l^{qs}$:
\begin{equation}
\widetilde {\rm{X}}_l^{qq}{\rm{  =  TTM(X}}_l^{qq}{\rm{,}}{{\rm{M}}_l^q}{\rm{)}},
\end{equation}
\begin{equation}
\overline {\rm{X}} _l^{qq}{\rm{  =  TTM(}}\widetilde {\rm{X}}_l^{qq} \oplus {{\rm{T}}_{l + 1}}{\rm{,}}{{\rm{M}}_l^q}{\rm{)}},
\end{equation}
Finally, we used a residual structure to update ${\rm{T}}$
\begin{equation}
{{\rm{T}}_l} = {{\rm{T}}_{l + 1}} + \overline {\rm{X}} _l^{qs} + \overline {\rm{X}} _l^{qq},
\label{eq:tl}
\end{equation}

Up to now, we have gotten all the output of TBTM, $\widehat {\rm{M}}_l^q$ and ${{\rm{T}}_l}$. Noting that we upsampled ${{\rm{T}}_l}$ before taking it as the input of the next TBTM to make sure the spatial sizes match.

\subsection{Target-aware Transformer Module}
The traditional Transformer has global attention, it is diffcult to make the model only focus on specific categories due to the support images contain multiple objects. Therefore, we propose the Target-aware Transformer Module (TTM) to make the model only calculates the hypercorrelation of the target in the mask. TTM consists of multiple Target-aware Transformer Layers(TTL), and the structure of TTL is illustrated in Fig. \ref{fig:TTL}. In order to gradually reduce support spatial sizes, \ie ${{H^s} \, {W^s}}$, we replaced the linear layer with a convolution layer to project input ${{\rm{X}}_{in}}$ into ${{\rm{X}}_Q},{{\rm{X}}_K},{{\rm{X}}_V}$, and a shortcut term ${{\rm{X}}_{SC}}$:
\begin{equation}
{\rm{X}}_\bigstar = {\rm{Conv}}_\bigstar({\rm{Drop(X}}_{in})),
\end{equation}
where $\bigstar \in \{Q,K,V,SC\}$, ${{\rm{X}}_{in}} \in \mathbb{R}^{{{H^q} \, {W^q} \times H^s \, W^s \times {D_{in}}}}$, and Drop means randomly setting elements 0 with rate $\beta$. We only perform drop operation on self-hypercorrelation branch, \ie ${\rm{X}}_{in}\!=\!{\rm{X}}_l^{qq}$. Taking the mask as a filter so that only foreground information is retained in the ${\rm{X}}_{V}$. We regarded query spatial sizes, \ie ${{H^q} \, {W^q}}$, as batchsize and carried out Batch Matrix Multiplication(BMM) to compute:
\begin{equation}
{\dot {\rm{X}}_{out}} = {\rm{Soft\max}} ({{\rm{X}}_Q}{X_K}^T){({\rm{X}}_V \odot {\widetilde M^s})},
\end{equation}
where:
\begin{equation}
{\widetilde M^s} = {{\rm{DownSample(M}}^s}) \in {\{ 0,1\} ^{ \ddot H^s \times  \ddot W^s}},
\end{equation}
and $\odot$ means broadcasted dot product. Two multi-layer perception and normalization layers follow to calculate the final output ${{\rm{X}}_{out}} \in {\mathbb{R}^{{H^q} \, {W^q} \times \dot H^s \, \dot W^s \times {D_{out}}}}$:
\begin{equation}
{{\rm \ddot {X}}_{out}} = {\rm{Norm(MLP}}({\dot X_{out}}) + {\dot X_{out}} + {X_{SC}}),
\label{eq:res}
\end{equation}
\begin{equation}
{{\rm {X}}_{out}} = {\rm{Norm(MLP}}({\ddot X_{out}}) + {\ddot X_{out}}),
\end{equation}
Now we have reduced support spatial sizes from $ H^s \, W^s$ to $ \dot H^s \, \dot W^s$ meanwhile changed channels from $D_{in}$ to $D_{out}$.

\subsection{Segmentation Decoder}
The structures of both decoders are the same as Convolution Block in the TBTM. It is simple but efficient to obtain the final prediction $\widehat {\rm{M}}_1^q \in \mathbb{R}^{2 \times H_1^q \times W_1^q}$.

The model parameters are optimized by the cross-entropy loss between a series of predictions $\{ \widehat {\rm{M}}_l^q\} _{l = 1}^4$ and the ground-truth ${{\rm{M}}^q} \in {\{ 0,1\} ^{H \times W}}$ overall pixel locations. Noting that we unsampled all the predictions to the same size with $M_q$ by bilinear interpolation before computing loss. We also set a hyperparameter $\alpha$ to adjust the weighs of $\{{\mathcal{L}_l}={\rm{CE}}(\widehat {\rm{M}}_l^q,{{\rm{M}}^q})\}_{l = 1}^4$:
\begin{equation}
{{\mathcal{L}}_{total}} = (1 - 3 \times \alpha ){\mathcal{L}_1} + \alpha \sum\limits_{l = 2}^4 {{\mathcal{L}_l}},
\end{equation}
where CE denotes cross-entropy and $\alpha=0.1$ in all the experiments.
\begin{figure}[ht]
	\begin{minipage}[ht]{0.49\linewidth}
	\newcommand{\s}{1}
	\begin{center}
	\resizebox{\linewidth}{!}{
	\begin{tikzpicture}[
		scale=\s,
		RoundRec/.style={rectangle,rounded corners=\s*1mm,draw,gray,very thick,minimum height =6mm, minimum width=10mm,},
		SquareNode/.style={rectangle,draw=green!50,fill=green!20,very thick, minimum size=5mm},
		ArrowC1/.style={rounded corners=\s*2mm,>={Latex[length = 2mm, width = 1.5mm,scale=\s]},line width=\s*1.5pt},
		]
	\tikzstyle{every node}=[scale=\s]
	
	\node [SquareNode] (Xin) at (0,0) {$\mathrm{X}_{in}$};
	\node [SquareNode] (Xout) at (0,9) {$\mathrm{X}_{out}$};
	\node [SquareNode,minimum size=6mm,draw=black!100,fill=white!20,font=\fontsize{7}{7}\selectfont] (mask) at (2.1,2) {$\mathrm{Mask}$};
	\node (odot) [font=\fontsize{7}{7}\selectfont,scale=2] at (3,2.7) {$\odot$};
	\node [RoundRec, fill=violet!20] (Convsc) at (-3,1) {$\mathrm{Conv}_{SC}$};
	\node [RoundRec, fill=magenta!20] (Convq) at (-1,1) {$\mathrm{Conv}_Q$};
	\node [RoundRec, fill=teal!20] (Convk) at (1,1) {$\mathrm{Conv}_K$};
	\node [RoundRec, fill=cyan!20] (Convv) at (3,1) {$\mathrm{Conv}_V$};
	\node [RoundRec, fill=pink!20,minimum width=28mm,] (bmm1) at (0,2) {$\mathrm{BMM}$};
	\node [RoundRec, fill=blue!20,minimum width=28mm,] (softmax) at (0,3) {$\mathrm{Softmax}$};
	\node [RoundRec, fill=pink!20,minimum width=50mm,] (bmm2) at (1.1,4) {$\mathrm{BMM}$};
	\node [RoundRec, fill=lime!20,minimum width=28mm,] (mlp1) at (0,5) {$\mathrm{MLP}$};
	\node [RoundRec, fill=lightgray!20,minimum width=28mm,] (add1) at (0,6) {$\mathrm{Add\&Norm}$};
	\node [RoundRec, fill=lime!20,minimum width=28mm,] (mlp2) at (0,7) {$\mathrm{MLP}$};
	\node [RoundRec, fill=lightgray!20,minimum width=28mm,] (add2) at (0,8) {$\mathrm{Add\&Norm}$};
	
	\draw [->,ArrowC1,black!100] (Xin) node[yshift=12pt,font=\fontsize{5}{5}\selectfont]{$H^sW^s \times {D_{in}}$} -|(Convsc);
	\draw [->,ArrowC1,black!100] (Xin)-|(Convq);
	\draw [->,ArrowC1,black!100] (Xin)-|(Convk);
	\draw [->,ArrowC1,black!100] (Xin)-|(Convv);
	\draw [->,ArrowC1,black!100] (Convq) node[above right,yshift=6pt,font=\fontsize{5}{5}\selectfont]{$\dot H^s \dot W^s \times {D_{hid}}$} -- ++(0,0.7);
	\draw [->,ArrowC1,black!100] (Convk) node[above right,yshift=6pt,font=\fontsize{5}{5}\selectfont]{$\ddot H^s \ddot W^s \times {D_{hid}}$} -- ++(0,0.7);
	\draw [->,ArrowC1,black!100] (bmm1) node[above right,yshift=6pt,font=\fontsize{5}{5}\selectfont]{$\dot H^s \dot W^s \times \ddot H^s \ddot W^s$} -- ++(0,0.7);
	\draw [->,ArrowC1,black!100] (softmax) node[above right,yshift=6pt,font=\fontsize{5}{5}\selectfont]{$\dot H^s \dot W^s \times \ddot H^s \ddot W^s$} -- ++(0,0.7);
	\draw [->,ArrowC1,black!100] (0,4.3) node[above right,yshift=-2pt,font=\fontsize{5}{5}\selectfont]{$\dot H^s \dot W^s \times {D_{out}}$} -- (mlp1);
	\draw [->,ArrowC1,black!100] ($(Convv)+(0,3.3)$)|- (add1);
	\draw [->,ArrowC1,black!100] (mlp1) node[above right,yshift=6pt,font=\fontsize{5}{5}\selectfont]{$\dot H^s \dot W^s \times {D_{out}}$} -- ++(0,0.7);
	\draw [->,ArrowC1,black!100] (add1) node[above right,yshift=6pt,font=\fontsize{5}{5}\selectfont]{$\dot H^s \dot W^s \times {D_{out}}$} -- ++(0,0.7);
	\draw [->,ArrowC1,black!100] (mlp2) node[above right,yshift=6pt,font=\fontsize{5}{5}\selectfont]{$\dot H^s \dot W^s \times {D_{out}}$} -- ++(0,0.7);
	\draw [->,ArrowC1,black!100] (add2) node[above right,yshift=6pt,font=\fontsize{5}{5}\selectfont]{$\dot H^s \dot W^s \times {D_{out}}$} -- (Xout);
	\draw [-,ArrowC1,black!100] (Convv) node[above right,yshift=6pt,font=\fontsize{5}{5}\selectfont]{$\ddot H^s \ddot W^s \times {D_{out}}$}   -- ($(odot)+(0,-0.2)$);
	\draw [->,ArrowC1,black!100] ($(odot)+(0,0.2)$) -- ($(Convv)+(0,2.7)$);
	\draw [->,ArrowC1,black!100] (mask) |- ($(odot)+(-0.2,0)$);
	\draw [->,ArrowC1,black!100] (Convsc) node[above right,yshift=6pt,font=\fontsize{5}{5}\selectfont]{$\dot H^s \dot W^s \times {D_{out}}$} |- (add1);
	\draw [rounded corners=\s*2mm] (0,6.5) -- ++(-2.0,0)[->,ArrowC1,black!100] |- (add2);
	\end{tikzpicture}
	}
	\end{center}
	\captionsetup{width=0.9\textwidth}
	\caption{Illustration of the proposed Target-aware Transformer Layer's calculation process.}
	\label{fig:TTL}
	\end{minipage}
	\begin{minipage}[ht]{0.49\linewidth}
		\begin{center}
		\resizebox{\linewidth}{!}{
		    \begin{tikzpicture}[scale=1]
		    \begin{axis}[
		            xlabel=Dropout rate $\beta$,
		            ylabel=mIoU,
		            y tick label style={/pgf/number format/.cd,%
		                set thousands separator={}},
		            width=3in,
		            height=3in,
		            legend style={at={(0.5,-0.2)},anchor=north}
		            ]
		    \addplot[smooth,mark=*,black] plot coordinates {
		            (0,67.4)
		            (0.05,68.3)
		            (0.1,68.0)
		            (0.2,68.0)
		            (0.3,67.5)
		            (0.5,67.1)
		            };
		    \end{axis}
		    \end{tikzpicture}
		}
		\end{center}
		\captionsetup{width=0.9\textwidth}
		\caption{Ablation study on the dropout rate.}
		\label{fig:drop}
	\end{minipage}
\vspace{-10pt}
\end{figure}

\section{Experiments}
In this section, we conducted extensive experiments on PASCAL-$5^i$\cite{AmirrezaShaban2017OneShotLF} and COCO-$20^i$\cite{KhoiNguyen2019FeatureWA} datasets which are prevalent in the few-shot segmentation field. And we use mIoU and FB-IoU as metrics to compare our results with recently excellent methods. Finally, we analyze the influence of each proposed module through extensive ablation experiments.
All experiments are implemented on PyTorch\cite{AdamPaszke2019PyTorchAI}. Following HSNet\cite{JuhongMin2021HypercorrelationSF}, we use Adam\cite{DiederikPKingma2014AdamAM} as the optimizer to update model parameters and the learning rate is set to 0.001. The batch size is set to 8 for all experiments. Both query and support images' spatial sizes are set to 400x400 without any data augmentation. Borrowed from ASNet\cite{DahyunKangIntegrativeFL}, we set $H_2^q,W_2^q = 50$, $H_2^s,W_2^s,H_3^s,W_3^s,H_3^q,W_3^q = 25$ and $H_4^q,W_4^q,H_4^s,W_4^s = 13$. Different from other methods\cite{JuhongMin2021HypercorrelationSF,DahyunKangIntegrativeFL,SunghwanHong2022CostAW}, our train epoch is only set to 50 for PASCAL-$5^i$ and 20 for COCO, which is much less than others.
\subsection{Datasets}
PASCAL-$5^i$ includes PASCAL VOC 2012\cite{MarkEveringham2015ThePV} and extended annotations from SDS\cite{BharathHariharan2014SimultaneousDA} datasets, which contain 20 object categories of images. All the images are evenly divided into 4 folds $i = \{ 0,...,3\}$, each fold contains 5 classes images $C_{test}^i = \{ 5 \times i,...,5 \times i + 4\}$ for testing and the rest 15 classes $C_{train}^i = \{ 0,...,19\} - C_{test}^i$ for training. Following \cite{ZhuotaoTian2020PriorGF}, we randomly sampled 1000 support-query pairs for testing.

COCO-$20^i$\cite{KhoiNguyen2019FeatureWA} is based on MSCOCO\cite{TsungYiLin2014MicrosoftCC}, which is much more difficult than PASCAL-$5^i$. We divided it into 4 folds as same as PASCAL-$5^i$, but each fold contains 60 and 20 categories images for training and testing respectively.
\subsection{Comparison with State-of-the-Arts}
As shown in Table \ref{tab:PASCAL}, \ref{tab:COCO}, we compared the performance of TBTNet and recently excellent approaches on PASCAL-$5^i$\cite{AmirrezaShaban2017OneShotLF} and COCO-$20^i$\cite{KhoiNguyen2019FeatureWA} respectively. Extensive experiments indicate that our model can achieve higher accuracy and shorter train time with fewer parameters.
\begin{table*}[t]
\begin{center}
\caption{Performance comparison on PASCAL-$5^i$\cite{AmirrezaShaban2017OneShotLF}. Best results in \textbf{bold}, and second best are \underline{underlined}.}
\label{tab:PASCAL}
\resizebox{\linewidth}{!}{
\begin{tabular}{c c|c c c c c c|c c c c c c|c c}
\toprule[1pt]
Backbone & \multirow{2}*{Methods} & \multicolumn{6}{c|}{1-shot} & \multicolumn{6}{c|}{5-shot} & learnable & train\\
network & & $5^0$ & $5^1$ & $5^2$ & $5^3$ & mean & FB-IoU & $5^0$ & $5^1$ & $5^2$ & $5^3$ & mean & FB-IoU & params & epoch\\
\midrule
\multirow{6}*{ResNet50} & PFENet\cite{ZhuotaoTian2020PriorGF} & 61.7 & 69.5 & 55.4 & 56.3 & 60.8 & 73.3 & 63.1 & 70.7 & 55.8 & 57.9 & 61.9 & 73.9 & 10.8M &200\\ 
 & HSNet\cite{JuhongMin2021HypercorrelationSF} 	& 64.3 & 70.7 & 60.3 & 60.5 & 64.0 & 76.7 & 70.3 & 73.2 & 67.4 & \underline{67.1} & 69.5 & \underline{80.6} & 2.6M & -\\ 
 & SSP\cite{QiFan2022SelfSupportFS} 	& {60.5} & 67.8 & \textbf{66.4} & {51.0} & {61.4} & - & {67.5} & 72.3 & \textbf{75.2} & {62.1} & {69.3} & - & 8.7M & -\\ 
 & VAT\cite{SunghwanHong2022CostAW} 		& \underline{67.6} & \underline{72.0} & 62.3 & 60.1 & 65.5 & \underline{77.8} & \textbf{72.4} & 73.6 & {68.6} & 65.7 & \underline{70.1} & \textbf{80.9} & 3.2M & 300\\ 
 & IPRNet\cite{AtsuroOkazawa2023InterclassPR} 	& 65.2 & \textbf{72.9} & \underline{63.3} & \underline{61.3} & \underline{65.7} & - & 70.2 & \textbf{75.6} & \underline{68.9} & 66.2 & \textbf{70.2} & - & - & 200\\ \cmidrule{2-16}
 & \textbf{Ours}& \textbf{68.7} & \underline{72.0} & {62.4} & \textbf{62.6} & \textbf{66.4} & \textbf{77.9} & \underline{70.6} & \underline{75.0} & 66.6 & \textbf{68.1} & \underline{70.1} & 80.1 & 0.3M & 50\\ \midrule
\multirow{6}*{ResNet101} & PFENet\cite{ZhuotaoTian2020PriorGF} & 60.5 & 69.4 & 54.4 & 55.9 & 60.1 & 72.9 & 62.8 & 70.4 & 54.9 & 57.6 & 61.4 & 73.5 & 10.8M &200\\ 
 & HSNet\cite{JuhongMin2021HypercorrelationSF} 	& 67.3 & 72.3 & \underline{62.0} & 63.1 & 66.2 & 77.6 & 71.8 & 74.4 & \underline{67.0} & 68.3 & 70.4 & 80.6 & 2.6M & -\\ 
 & ASNet\cite{DahyunKangIntegrativeFL} 	& 69.0 & 73.1 & \underline{62.0} & \underline{63.6} & \underline{66.9} & \underline{78.0} & \underline{73.1} & 75.6 & 65.7 & \underline{69.9} & \underline{71.1} & \underline{81.0} & 1.3M & 500\\  
 & IPMT\cite{YuanweiLiu2022IntermediatePM} 	& \textbf{71.6} & \textbf{73.5} & 58.0 & 61.2 & 66.1 & - & \textbf{75.3} & \textbf{76.9} & 59.6 & 65.1 & 69.2 & - & - & 200\\ \cmidrule{2-16}
 & \textbf{Ours} & \underline{70.2} & \underline{73.3} & \textbf{63.6} & \textbf{66.1} & \textbf{68.3} & \textbf{79.0} & 72.2 & \underline{76.0} & \textbf{68.3} & \textbf{71.5} & \textbf{72.0} & \textbf{81.6} & 0.4M & 50\\ \bottomrule[1pt]
\end{tabular}
}
\end{center}
\vspace{-10pt}
\end{table*}
\begin{table*}[t]
\begin{center}
\caption{Performance comparison on COCO-$20^i$\cite{KhoiNguyen2019FeatureWA}.}
\label{tab:COCO}
\resizebox{\linewidth}{!}{
\begin{tabular}{c c|c c c c c c|c c c c c c|c c}
\toprule[1pt]
Backbone & \multirow{2}*{Methods} & \multicolumn{6}{c|}{1-shot} & \multicolumn{6}{c|}{5-shot} & learnable & train\\
network & & $5^0$ & $5^1$ & $5^2$ & $5^3$ & mean & FB-IoU & $5^0$ & $5^1$ & $5^2$ & $5^3$ & mean & FB-IoU & params & epoch\\
\midrule
\multirow{5}*{ResNet50} & PFENet\cite{ZhuotaoTian2020PriorGF} & 36.5 & 38.6 & 34.5 & 33.8 & 35.8 & - & 36.5 & 43.3 & 37.8 & 38.4 & 39.0 & - & 10.8M & 50\\ 
 & CMNet\cite{98905c56b820287462fded935c888d24db88cc7b} 	& \textbf{48.7} & 33.3 & 26.8 & 31.2 & 35.0 & - & \textbf{49.5} & 35.6 & 31.8 & 33.1 & 37.5 & - & - & 50\\ 
 & IPMT\cite{YuanweiLiu2022IntermediatePM} 	& \underline{41.4} & \underline{45.1} & \textbf{45.6} & \underline{40.0} & \underline{43.0} & - & {43.5} & 49.7 & 48.7 & \textbf{47.9} & {47.5} & - & - & 50\\ 
 & VAT\cite{SunghwanHong2022CostAW} 		& 39.0 & 43.8 & 42.6 & 39.7 & 41.3 & \underline{68.8} & 44.1 & \underline{51.1} & \underline{50.2} & 46.1 & \underline{47.9} & \underline{72.4} & 3.3M & -\\ \cmidrule{2-16}
 & \textbf{Ours}& {39.8} & \textbf{46.9} & \underline{44.6} & \textbf{43.8} & \textbf{43.8} & \textbf{70.6} & \underline{45.6} & \textbf{54.7} & \textbf{51.5} & \underline{47.2} & \textbf{49.7} & \textbf{72.7} & 0.3M & 20\\ \midrule
\multirow{5}*{ResNet101} & PFENet\cite{ZhuotaoTian2020PriorGF} & 34.3 & 33.0 & 32.3 & 30.1 & 32.4 & - & 38.5 & 38.6 & 38.2 & 34.3 & 37.4 & - & 10.8M & 50\\ 
 & HSNet\cite{JuhongMin2021HypercorrelationSF} 	& 37.2 & 44.1 & 42.4 & 41.3 & 41.2 & 69.1 & 45.9 & \underline{53.0} & \underline{51.8} & 47.1 & \underline{49.5} & 72.4 & 2.6M & -\\ 
 & ASNet\cite{DahyunKangIntegrativeFL} 	& \textbf{41.8} & 45.4 & 43.2 & \underline{41.9} & \underline{43.1} & \underline{69.4} & \textbf{48.0} & 52.1 & 49.7 & \underline{48.2} & \underline{49.5} & \underline{72.7} & 1.3M & -\\ 
 & IPMT\cite{YuanweiLiu2022IntermediatePM} 	& \underline{40.5} & \underline{45.7} & \underline{44.8} & 39.3 & 42.6 & - & 45.1 & 50.3 & 49.3 & 46.8 & 47.9 & - & - & 50\\ \cmidrule{2-16}
 & \textbf{Ours} & 40.2 & \textbf{47.5} & \textbf{46.6} & \textbf{45.3} & \textbf{44.9} & \textbf{71.2} & \underline{46.2} & \textbf{55.5} & \textbf{52.7} & \textbf{49.4} & \textbf{50.9} & \textbf{73.3} & 0.4M & 20\\ \bottomrule[1pt]
\end{tabular}
}
\end{center}
\vspace{-10pt}
\end{table*}
TBTNet outperformed all other models on PASCAL-$5^i$ whether took ResNet50 or ResNet101 as the backbone. It achieved the best or second-best results on each fold, especially over ASNet 2.5 mIoU on fold 3 with ResNet101. TBTNet exceeds the previous SOTA model ASNet 1.4 mIoU and achieved a new record. As for the numbers of learnable parameters, our TBTNet only has 0.4M which is 3.7\% of PFENet’s and 30.8\% of ASNet’s. Due to the small number of parameters, our model is easy to train and only needs 50 epochs to converge which is 10\% of ASNet’s and 25\% of others. To the best of our knowledge, TBTNet is the model with the shortest training period to date.

On the more difficult datasets COCO-$20^i$, TBTNet also achieved remarkable performance. Our model got the best score on folds 1, 2, 3 and mIoU, no matter whether in 1-shot or 5-shot conditions and two backbones. It manifests that TBTNet can generalize well, with almost no bias towards categories. Under 1-shot configuration, TBTNet outperformed ASNet by 1.8 mIoU when taking ResNet101 as backbone. As on the PASCAL dataset, our training period was only 40\% of the others.
\begin{figure*}[ht]
\begin{center}
\resizebox{\textwidth}{!}{
\begin{tikzpicture}[scale=1,]
\tikzstyle{every node}=[scale=1]
	\foreach \y/\idx/\f in {1/28/3,3/50/0,5/74/0,7/103/3}{
		\foreach \x/\n in {1/2,3/3,5/10,7/11,9/12,11/13,13/5,15/15} {\node at (\x,\y){\includegraphics[width=19.5 mm]{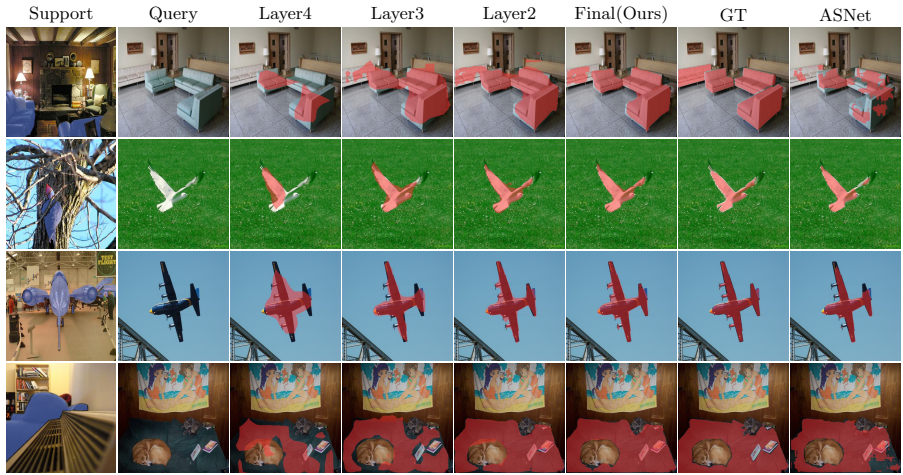}};}
	}
	\foreach \x/\z in {1/Support,3/Query,5/Layer4,7/Layer3,9/Layer2,11/Final(Ours),13/GT,15/ASNet} {\node at (\x,8.2){\z};}
		
\end{tikzpicture}
}
\end{center}
\caption{Qualitative comparison between our proposed TBTNet and ASNet. From left to right: support image, query image, intermediate prediction of TBTNet at layer 4, 3, 2, finale prediction, ground truth and the prediction of ASNet.}
\label{fig:qualitative}
\vspace{-10pt}
\end{figure*}

In Fig. \ref{fig:qualitative}, we visualized the inference procedure of TBTNet and compared the predictions with ASNet, which is one of the SOTA models. We observed that segmentation can be optimized gradually with increasing resolution layer by layer. Our model can use self-similarity to make the segmentation more complete.
\subsection{Ablation Study}
\label{AblationStudy}
All ablation study experiments are carried out on PASCAL-$5^i$\cite{AmirrezaShaban2017OneShotLF} with the ResNet101 backbone and the 1-shot setting.

\subsubsection{Effectiveness of Bi-Transformer architecture.} We conduct an ablation study by modifying the structure of TBTM to evaluate the influence of self-similarity. As shown in the Table \ref{tab:ablation}, \enquote{Bi-T} means whether utilize self-similarity branch in the TBTM. In other words, the unselected \enquote{Bi-T} represents  ${{\rm{T}}_l} = {{\rm{T}}_{l + 1}} + \overline {\rm{X}} _l^{qs}$, and vice versa ${{\rm{T}}_l} = {{\rm{T}}_{l + 1}} + \overline {\rm{X}} _l^{qs} + \overline {\rm{X}} _l^{qq}$ in Eq.(\ref{eq:tl}). Experiments indicate that self-similarity branch can lead to 3.1\% increase in mIoU. The improvement can prove that Bi-Transformer architecture is very efficacious for FSS.

\subsubsection{Effectiveness of TTL.} To explore the strength of the proposed TTL, we compare it with Attention Squeeze Layer (ASL)\cite{DahyunKangIntegrativeFL}. In the Table \ref{tab:ablation}, \enquote{TTL} and \enquote{ASL} denote the sub-module in TTM. We set $\beta$ of both experiments as 0.05 for fairness. When TTL is replaced with ASL, a significant drop can be observed, with mIoU descending from 68.3 to 67.7. It indicates that our proposed TTL is more efficient than ASL, which may benefit from a more multivariate residual structure in Eq.(\ref{eq:res}).

\subsubsection{Ablation study on the dropout rate.} We conducted a series of experiments to find the optimal parameter $\beta$, and all the results are shown in Fig. \ref{fig:drop}. The mIoU reaches its peak at 68.3 when $\beta$ is 0.05. As $\beta$ increases, mIoU rises and then falls. It is because appropriate $\beta$ can effectively prevent overfitting, and enhance the generalization ability of the model, whereas an excessive $\beta$ will lead to the loss of too much information, thus hindering performance.
\begin{table}[t]
\begin{center}
\caption{Ablation study on the Bi-Transformer architecture and our proposed TTL.}
\label{tab:ablation}
\resizebox{0.5\linewidth}{!}{
\begin{tabular}{c c c c|c}
\toprule[1pt]
Bi-T & TTL & ASL & Dropout rate & mIoU\\ \midrule
 & \checkmark &   & 0 & 65.4({\color{red}+0.0}) \\ 
\checkmark & \checkmark &   & 0 & 67.4({\color{red}+2.0}) \\ 
\checkmark & \checkmark &  & 0.05 & 68.3({\color{blue}+0.0}) \\ 
\checkmark &  & \checkmark & 0.05 & 67.7({\color{blue}-0.6}) \\ \bottomrule[1pt]
\end{tabular}
}
\end{center}
\vspace{-10pt}
\end{table}
\section{Conclution}
In this paper, we introduce Bi-Transformer architecture to few-shot segmentation. To utilize self-similarity information efficiently, we proposed TBTM to integrate it with cross-similarity. A novel TTL is also been proposed to compact the similarity information which is a variant of the transformer. Our TBTNet is a lightweight and fast convergence model. Its effectiveness has been demonstrated by its outstanding performance on the standard benchmarks for few-shot segmentation. We hope that our research will shed light on other domains where similarity analysis is required.
%
%
%
\bibliographystyle{splncs04}
\bibliography{export}
%

\end{document}